\documentclass{article}
\usepackage{ijcai22}

\usepackage{times}
\usepackage{soul}
\usepackage{url}
\usepackage[hidelinks]{hyperref}
\usepackage[utf8]{inputenc}
\usepackage[small]{caption}
\usepackage{graphicx}
\usepackage{amsmath}
\usepackage{amsthm}
\usepackage{booktabs}
\usepackage{algorithm}
\usepackage{algorithmic}
\title{Integration of knowledge and data in machine learning}

\author{
Yuntian Chen$^1$\And
Dongxiao Zhang$^2$$^,$$^3$$^,$\footnote{Contact Author}\\
\affiliations
$^1$Eastern Institute for Advanced Study, Yongriver Institute of Technology, Zhejiang, P. R. China\\
$^2$National Center for Applied Mathematics Shenzhen (NCAMS), Southern University of Science and Technology, Guangdong, P. R. China\\
$^3$Department of Mathematics and Theories, Peng Cheng Laboratory, Guangdong, P. R. China
\emails
\
ychen@eias.ac.cn,
zhangdx@sustech.edu.cn
}

\begin{document}

\maketitle

\begin{abstract}
Scientific research’s mandate is to comprehend and explore the world, as well as to improve it based on experience and knowledge. Knowledge embedding and knowledge discovery are two significant methods of integrating knowledge and data. Through knowledge embedding, the barriers between knowledge and data can be eliminated, and machine learning models with physical common sense can be established. Meanwhile, humans’ understanding of the world is always limited, and knowledge discovery takes advantage of machine learning to extract new knowledge from observations. Knowledge discovery can not only assist researchers to better grasp the nature of physics, but it can also support them in conducting knowledge embedding research. A closed loop of knowledge generation and usage are formed by combining knowledge embedding with knowledge discovery, which can improve the robustness and accuracy of models and uncover previously unknown scientific principles. This study summarizes and analyzes extant literature, as well as identifies research gaps and future opportunities.
\end{abstract}

\section{Introduction}

The primary objective of scientific research is to understand and investigate the world, as well as to improve it based on experience and knowledge. Scientific advancement is frequently marked by an alternation of science and engineering growth. On the one hand, it constitutes the exploration and discovery of new mechanisms via practice and experiment, as well as the deepening of knowledge of the physical world (i.e., scientific development); on the other hand, it is the application of existing knowledge to practice (i.e., engineering progress). 

Engineering practice is guided by knowledge, and the data gathered in practice, in turn, contribute to the advancement of science. In the 16th century, for example, Tycho Brahe established an observatory and accumulated a vast amount of observation data, based on which Johannes Kepler proposed Kepler's law, and Isaac Newton derived the law of gravity from it. The gained knowledge could then be used for the development of new observational equipment, such as the Harper Telescope.

As technology progresses, researchers are able to collect an increasing number of observations. This has led to the widespread use of machine learning as a statistical modeling tool with powerful fitting capabilities in various fields. In science, machine learning can inspire scientists to find new knowledge \cite{AIguide}, and even deduce basic theorems \cite{kaliszyk2018reinforcement}. In engineering, machine learning, as opposed to classic mechanism-based simulations, can predict changes in physical fields using data-driven methods. Nevertheless, it still faces the problem of low accuracy and robustness caused by data scarcity and complex scenarios. Indeed, it is difficult to obtain the desired performance by simply applying machine learning directly. Embedding domain knowledge to provide richer information for models, however, is a practical way to improve model performance \cite{karpatne2017theory}.

\begin{figure*}
    \centering
    \includegraphics[width=15cm]{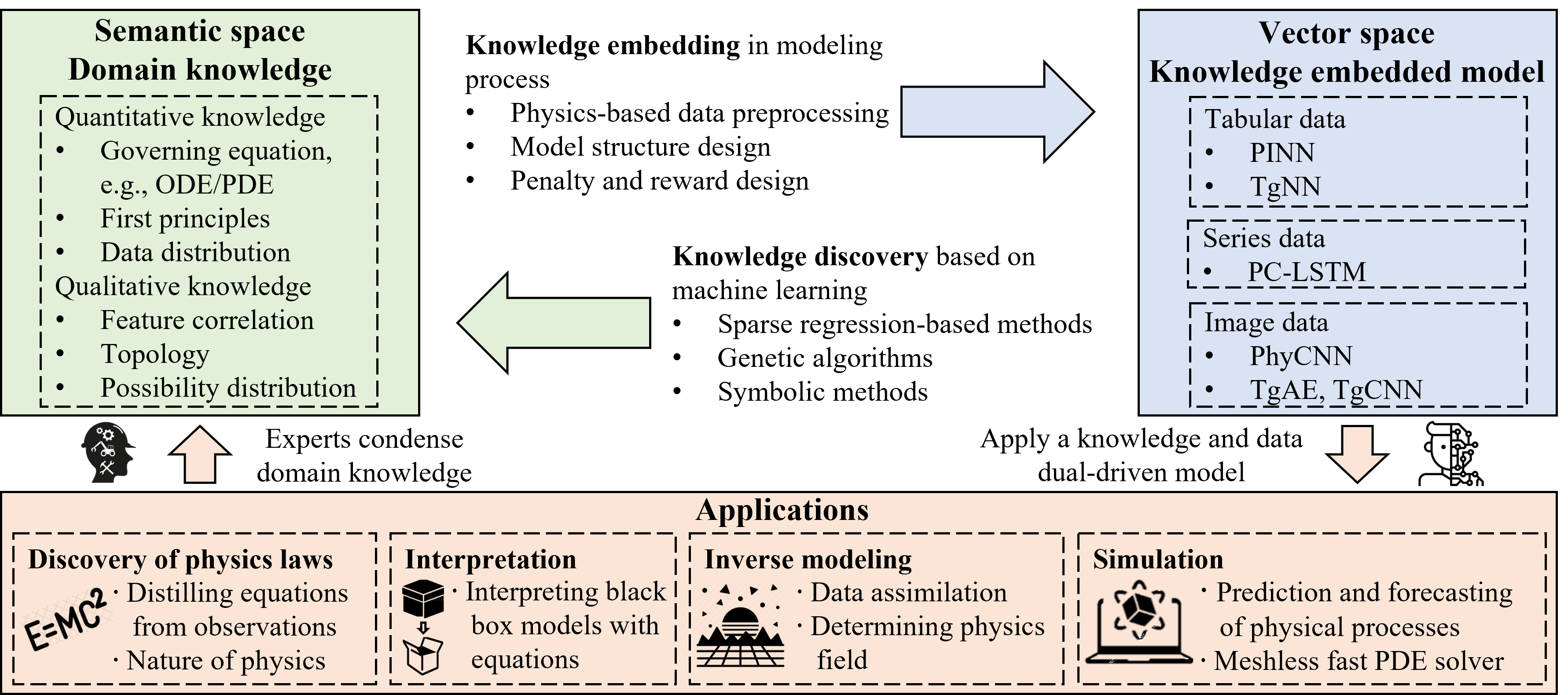}
    \caption{Schematic diagram of the relationship between knowledge embedding and knowledge discovery}
\end{figure*}

Researchers’ attempts to integrate domain knowledge with data-driven models may be generally divided into two categories: knowledge embedding and knowledge discovery. {\it Knowledge embedding} is the process of incorporating domain knowledge into data-driven models in order to create models that have physical common sense, improve model accuracy and robustness, reduce data requirements, and create land-ready machine learning models. {\it Knowledge discovery} is the process of directly mining governing equations from observations and experimental data through machine learning algorithms, and inspire scientific study.

Knowledge embedding and knowledge discovery are intertwined, and they can form a closed loop. A schematic diagram of the system is shown in Figure 1. On the one hand, by using domain knowledge obtained from expert experience and theoretical study in the semantic space, knowledge embedding can improve machine learning models in the vector space (blue arrow in Figure 1). On the other hand, because the systems are disordered and irregular in many practical applications, the structure and coefficients of potential governing equations are often too complex to obtain from theoretical derivations. Knowledge discovery can condense domain knowledge from data to support knowledge embedding (green arrow in Figure 1).

Regarding applications, knowledge embedding can improve the performance of machine learning models and facilitate the development of efficient simulators and inverse modeling. Knowledge discovery, however, can be used to discover new physical principles, as well as to provide interpretability for black-box models. Knowledge embedding and knowledge discovery are the key issues in achieving the integration of domain knowledge and machine learning algorithms. In the last decade, researchers have produced a large body of pertinent exploratory work. This paper aims to outline current studies in knowledge discovery and knowledge embedding, and provide insights into research gaps and future opportunities.

\section{Knowledge Discovery}

The goal of knowledge discovery is to extract undiscovered knowledge from data and push the boundaries of human intelligence forward. In early scientific work, researchers obtained equation structures by theoretical derivation and then determined the coefficients via regression methods \cite{hosmer2013applied}, such as in the discovery of the law of gravity and Maxwell’s equations. Because many real-world problems, such as turbulence in fluid dynamics, are too complicated to be solved using first-principle models, researchers have developed simulation approaches \cite{griebel1998numerical,zhang2001stochastic}. However, simulations fail to reveal the full internal structure of complex systems and lack interpretability \cite{ARE}. With the development of machine learning, neural networks are utilized as approximators to handle knowledge discovery problems, such as DeepONet \cite{DeepONet}. Although theory demonstrates that the neural network can approximate any function and its derivative \cite{hornik1990universal}, its essence is a surrogate model (i.e., unexplainable black box), and no explicit knowledge is obtained. Researchers have also attempted to use the physics-informed neural network (PINN) to determine the governing equations \cite{PINN}, but such approach requires the explicit form of the governing equation, which essentially constitutes an inverse problem and not knowledge discovery.

The real knowledge discovery method is capable of directly extracting the governing equation that best matches the data with transfer ability when the equation structure is unknown. The core of knowledge discovery is determining the structure and coefficients of the governing equation. The complexity of the equation structure is the first criterion for evaluating knowledge discovery methods. The second evaluation dimension is the complexity of the equation coefficients (Figure 2).

\subsection{Mining equations with complex structures}

Existing knowledge discovery methods will first construct candidate sets (i.e., library of function terms) based on prior knowledge, and then choose the most appropriate combination of candidate terms to produce equations through various optimization methods. Existing methods can be divided into closed library methods, expandable library methods, and open-form equation methods, which are shown in Figure 2.

Closed library methods are the most widely used, and are based on sparse regression for distilling the dominating candidate function terms. \citeauthor{SINDy} [\citeyear{SINDy}] developed SINDy, which built an overcomplete library first and then utilized sparse regression to determine the appropriate equation. Although SINDy only deals with ordinary differential equation (ODE), it has many variants \cite{champion2019discovery}. For example, PDE-FIND extends SINDy to partial differential equation (PDE) by introducing partial derivative terms in the library \cite{PDEFIND}. SGTR combines group sparse coding and solves the problem of parametric PDEs \cite{SGTR}. Moreover, different norm minimizations as sparsity constraints can be used in sparse regression algorithms \cite{donoho2003optimally,hoyer2004non}. For noisy observations in practice, high quality data can be generated by low-rank denoising \cite{DLrSR} and neural network fitting \cite{PeRCNN,DLGA}. In addition to selecting candidate terms, closed library methods can also be used to automatically determine physical processes \cite{chang2019identification}, which deepens our understanding of the nature of physics.

Since the candidate sets of closed library methods are preset, prior information can be easily embedded. For instance, \citeauthor{PeRCNN} [\citeyear{PeRCNN}] utilize specially-designed kernels to encode known terms. Especially in PDE-Net, each $\delta$t block corresponds to a time step, which establishes the connection between the governing equation and the network \cite{PDENET}. There are also numerous variants of PDE-Net, most of which rely on the preset overcomplete library. Furthermore, PINN-SR is proposed by combining PINN with sparse regression to embed domain knowledge into the knowledge discovery model \cite{PINNSR}.

Although closed library methods based on sparse regression are easy to implement, they fall into dilemma in practice. Specifically, conventional approaches can identify most of the governing equations of simple systems, but it is difficult to provide an overcomplete candidate set for complex systems that cannot be solved by conventional methods. Therefore, the expandable library is more suitable for discovering governing equations with complex structures than the closed library (Figure 3b). \citeauthor{EPDE} [\citeyear{EPDE}] proposed EPDE to verify the impact of genetic algorithms in PDE discovery. Then, DLGA integrates the neural network and the genetic algorithm, and realizes automatic expansion of the candidate set by encoding different function terms as gene segments \cite{DLGA}. The variants of DLGA have explored knowledge discovery under noisy and scarce data, and especially R-DLGA obtained high robustness by combining PINN \cite{RDLGA}. In addition to genetic algorithms, PDE-Net 2.0 \cite{PDENET2} introduces SymNet \cite{sahoo2018learning}, which uses network topology to produce interaction terms. Nevertheless, both PDE-Net 2.0 and genetic algorithms can only generate new function terms through addition and multiplication, and cannot implement division operations or generate composite functions. As a consequence, despite the fact that expandable library methods are more flexible and use less memory than closed library methods\cite{PDENET2}, they are still unable to construct governing equations with fractional structures and compound functions.

\begin{figure}
    \centering
    \includegraphics[width=8cm]{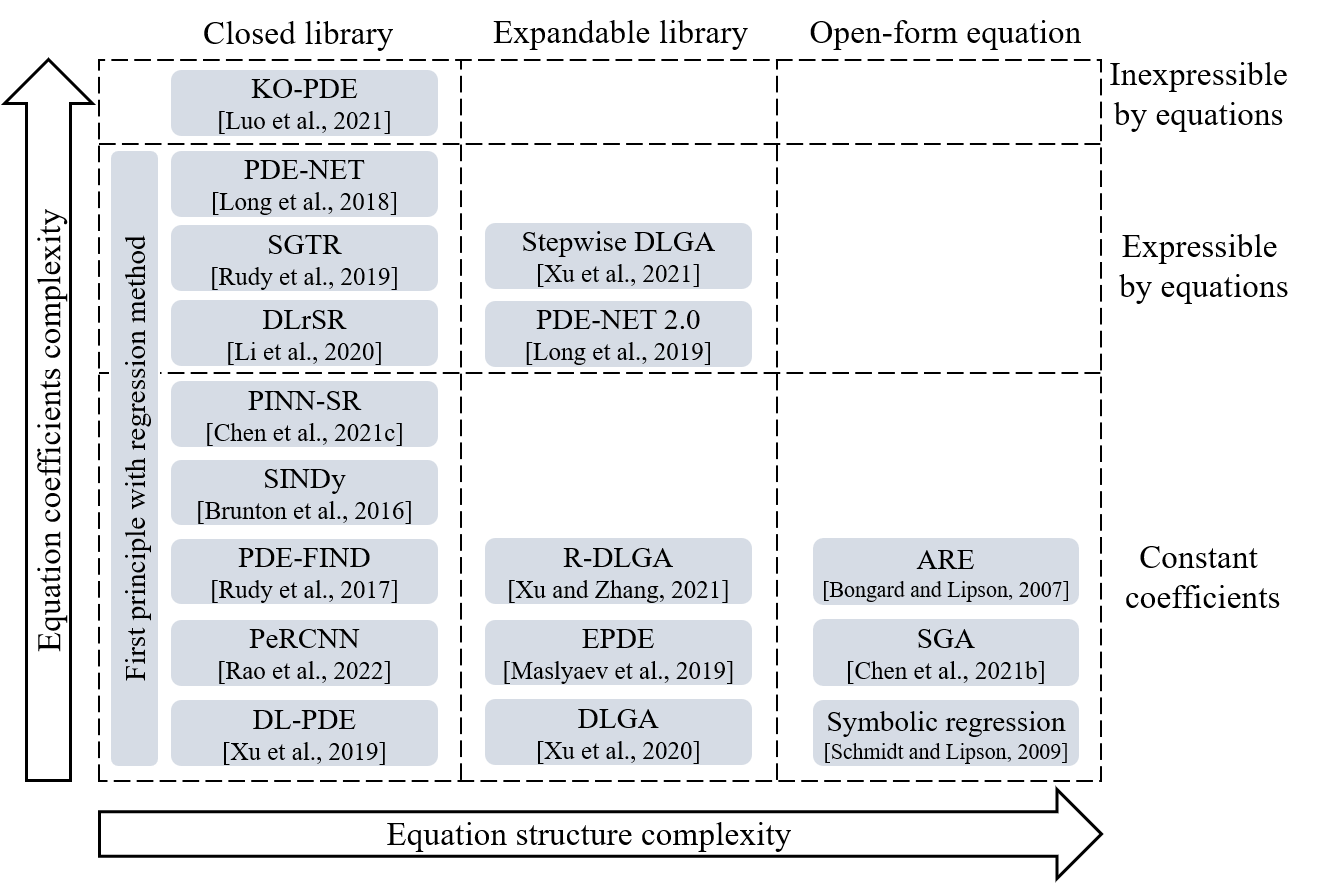}
    \caption{Diagram of the classification of knowledge discovery algorithms.}
\end{figure}

\begin{figure}
    \centering
    \includegraphics[width=7cm]{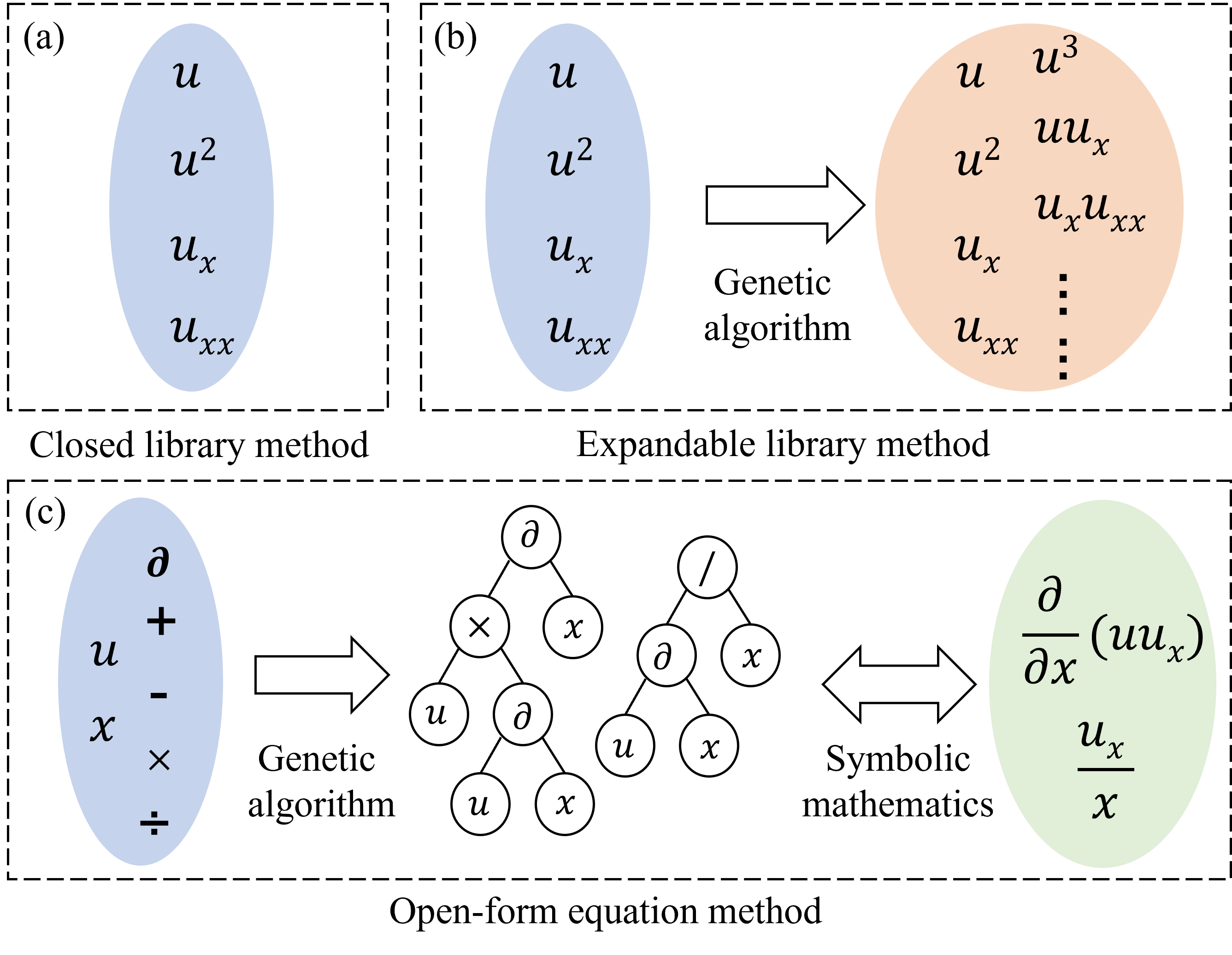}
    \caption{Illustration of closed library methods, expandable library methods, and open-form equation methods.}
\end{figure}

In order to mine arbitrary equations from data, open-form equation methods are proposed, as shown in Figure 3c. For instance, automated reverse engineering automatically generates equations for a nonlinear coupled dynamical system with the assistance of symbolic mathematics \cite{ARE}. However, because this method examines each variable separately, there are scalability issues. Later, researchers conducted more work on symbolic regression, and recommended that the governing equation be represented by binary trees (Figure 3c) \cite{schmidt2009distilling}. The above methods, however, are prone to overfitting and can only produce simple polynomials or ODEs \cite{PeRCNN}. SGA provides a tree-based genetic algorithm that can handle increasingly complicated systems and accomplish PDE discovery using symbolic mathematical representation \cite{SGA}. Due to the wider optimization space of open-form equation methods, they have greater computational cost than conventional methods in practice.

The three above-mentioned methods are applicable to different scenarios. If the system under study is simple, closed library methods, such as sparse regression, are both accurate and efficient. Expandable library methods, such as genetic algorithms, are more appropriate for systems with complicated interaction terms and have a low memory requirement. For a strongly nonlinear system with multi-physics coupling, governing equations may be highly complex, and a larger exploration space can be generated with the assistance of symbolic mathematics to realize optimization of arbitrary open-form equations.

\subsection{Mining equations with complex coefficients}

When mining governing equations from data, the complexity of the coefficients is significant. The coefficients can be divided into three groups, as illustrated in Figure 4, i.e., constant coefficients, coefficients that are expressible by equations, and coefficients that are inexpressible by equations. It is worth mentioning that the inexpressible coefficient field in Figure 4c is generated by Karhunen-Loeve expansion (KLE) with 20 random variables, which is a commonly used random field generation method in simulations. 

The method of mining constant coefficient equations is straightforward. After obtaining the equation structure, the least squares approach is all that is required to fit the coefficients. Therefore, all of the methods described in the preceding section can handle constant coefficients.

In realistic circumstances, there are many parametric governing equations. Their coefficients will vary over time or space, and can be described by equations, such as trigonometric functions, as shown in Figure 4b. The challenge of such problems is that the structure and coefficients of the equation are all unknown, and the optimal equation structure may be different in different coefficient value intervals. When the coefficients of the equation change, the method is prone to overfitting to local data, making finding the correct global solution difficult. In many studies, the equation structure is determined first through sparse regression, and then the coefficients are obtained through piecewise fitting \cite{SGTR} or pointwise fitting \cite{DLrSR}. The essence of the fitting method is to visualize the change trends of the coefficients, and the specific expression of the coefficients cannot be obtained. Stepwise-DLGA presents a winner-takes-all strategy, which groups the observations and chooses the frequent terms in distinct groups as the final result \cite{stepwiseDLGA}. Although the calculation process of Stepwise-DLGA is complex, it can avoid local overfitting and provide specific expressions of the coefficients. 

Many investigations divide the range of values into fitting windows, and then fit coefficients with constants within each window. However, when the coefficient field possesses strong nonlinearity, the assumption of constant coefficient is difficult to hold for large windows, and numerous overfitting equation structures will exist for narrow windows. As a result, the approaches described above can only solve the variable coefficient problem with weak nonlinearity (i.e., expressible by equations). In practice, however, many of the coefficients correspond to physical fields, resulting in significant nonlinearities (e.g., the permeability field in pollutant diffusion problems and the thermal conductivity field in heat transfer problems). In numerical simulations, since it is challenging to formulate such coefficient fields directly, they are even described by two-dimensional random fields \cite{zhang2001stochastic}, such as the coefficient field in Figure 4c. The kernel smoothing approach is used by KO-PDE to add nonlinearity in each fitting window \cite{KOPDE}. It attempts to avoid local overfitting by allowing the window to encompass as much nearby data as possible without destroying the nonlinearity of the coefficients in the window. The governing equation mining problem of complex coefficient fields is critical for knowledge discovery applications and requires further studies.

\begin{figure}
    \centering
    \includegraphics[width=8cm]{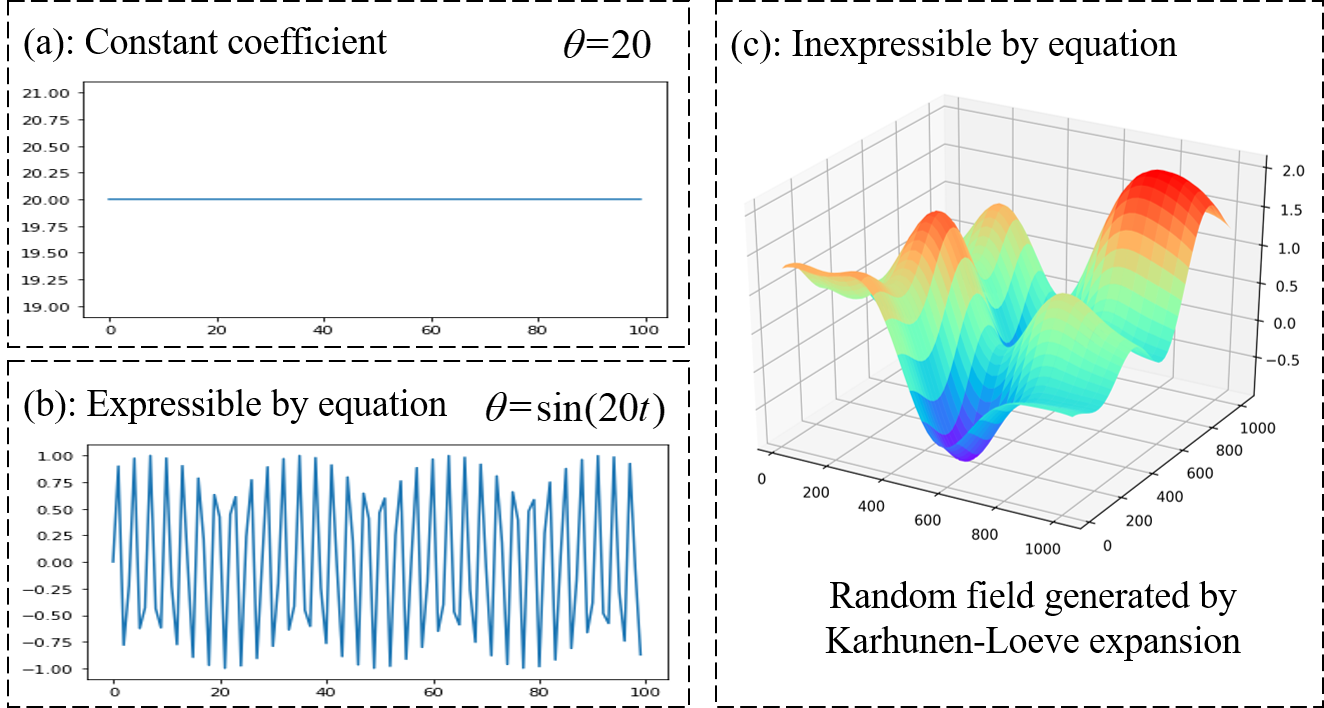}
    \caption{Illustration of different kinds of coefficients. (a): constant coefficient with a value of 20; (b): coefficient that can be described by trigonometric functions; (c): inexpressible random field that is used to describe physical fields, such as permeability fields.}
\end{figure}

\subsection{Challenges of knowledge discovery}

The representation of equations is the core issue in knowledge discovery. Closed library methods directly include all possible terms; although they are easy to implement, they have a restricted range of applications. In expandable library methods, the representation of PDEs is realized by representing the function terms as gene segments or kernels, so that the algorithm can find equations with complex interaction terms. Open-form equation methods, which can deal with governing equations with fractional structures and compound functions, employ symbolic mathematics to represent any form of governing equation, but the computational cost is high. In the future, more efficient and comprehensive equation representation approaches should be investigated.

There are five research gaps and future opportunities in knowledge discovery, including:

\begin{itemize}
\item In order to optimize equations via efficient gradient-based methods, a more appropriate embedding approach for equations is required (similar to the word vector \cite{le2014distributed}). The edit distance does not infer performance in equations (e.g., if the fourth derivative is the solution, the third derivative is not necessarily better than the second derivative).
\item Governing equations are essentially necessary conditions, but sufficient conditions are found in many cases, which leads to overfitting. Future studies might look towards discovering equations from multiple experiments \cite{tod2021discovering} to extract commonalities (i.e., necessary conditions).
\item Governing equations for complex systems, such as turbulence, are not only complex, but can even constitute a set of PDEs. Algorithms for mining equations with complex coefficients and structures are required (top right corner of Figure 2).
\item The precision of derivatives is important for mining PDEs. Gradients calculated by difference are not robust to noise \cite{PDEFIND}. Anti-noise methods include utilizing automatic differentiation in neural networks, using neural networks to generate high-quality data \cite{PeRCNN,DLGA}, and applying weak forms of equations \cite{xu2021deep}. PINN-SR and R-DLGA prove that robustness can be improved by embedding domain knowledge, which is worth exploring in the future.
\item The goal of knowledge discovery is to find an optimal balance between accuracy and parsimony of equations. As the library goes from closed to open, it is a process of gaining precision while diminishing simplicity. Open-form equation methods make it easy to find equivalent forms of equations. Determination of how to best simplify the equations presents a major challenge.

\end{itemize}

\section{Knowledge Embedding}

The knowledge, experience and physical mechanisms accumulated by human beings are valuable assets, but most current machine learning models fail to properly exploit them, which greatly limits the application of machine learning. Pure data-driven models not only have high data requirements, but might also produce predictions that violate the physical mechanism \cite{PINN,TgNN}. By integrating domain knowledge in machine learning models, it is possible to transcend the barriers between data-driven and knowledge-driven models.

\subsection{Knowledge embedding in the modeling process}

Researchers attempt to embed domain knowledge into the machine learning modeling process, including the following three steps: data preprocessing; model structure design; and penalty and reward design (Figure 5).

In the data preprocessing step, in addition to conventional feature engineering methods, domain knowledge can be applied to data normalization. For example, when assessing underground resources, the undulations of the formations are utilized as domain knowledge in the formation-adjusted stratified normalization to ensure that the strata of different wells remain aligned \cite{chen2020physics}. In biological research, the remove unwanted variation (RUV), constructed based on factor analysis on control genes, works better than conventional normalization for RNA-seq data \cite{risso2014normalization}. Moreover, time series data can also be decomposed using domain knowledge, such as in the forecasting of electrical load, which can be decomposed into inherent patterns related to forecast region, and influencing factors (e.g., weather conditions) pertinent to the particular forecast time \cite{chen2021theory}.

In model structure design, four embedding methods exist, as shown in Figure 5. Firstly, the network topology can be designed according to prior knowledge. Early research focused on computer vision. For example, researchers developed two-stream convolutional networks for action recognition, based on the human visual cortex’s two pathways \cite{simonyan2014two}. Researchers also improved computational saliency models based on biological visual salience detection \cite{yohanandan2018saliency}. In geoscience, \citeauthor{chen2020physics} [\citeyear{chen2020physics}] proposed a mechanism-mimic network architecture based on geomechanical equations. In addition, the structure of the $\delta$t block of PDE-Net is also determined according to temporal discretization \cite{PDENET}.

The second approach to embed domain knowledge in the model structure is to use the relationship between differentiation and convolution to design kernels \cite{PDENET,PDENET2}. For instance, in physics-constrained deep learning, the Sobel filter is used to calculate derivatives in CNN \cite{zhu2019physics}. In FEA-Net, the kernel is constructed according to finite element analysis (FEA), and the network is constructed based on the Jacobi solver \cite{yao2019fea}. In PeRCNN, the kernels in the model are used to represent gradients to generate high resolution data \cite{PeRCNN}.

The third approach is to design a neural network according to the finite element method (FEM), which converts the equations into a network. For example, \citeauthor{ramuhalli2005finite} [\citeyear{ramuhalli2005finite}] constructed finite-element neural networks by using unknown variables in the equation as weights in the network.

The fourth approach is to embed prior knowledge by constraining the value space of the model outputs. For example, \citeauthor{HCP} [\citeyear{HCP}] proposed hard constraint projection (HCP) to construct a projection matrix that maps the predictions of the neural network to a space that satisfies physical constraints, which can be regarded as a special activation function. PC-LSTM adds a ReLU function at the end of the network to ensure non-negativity of the outputs \cite{luo2021deep}. In computer vision, \citeauthor{pathak2015constrained} [\citeyear{pathak2015constrained}] proposed a two-step mapping method to embed domain knowledge and guarantee that the model outputs satisfy logical rules.

In penalty and reward design, domain knowledge is mainly transformed into constraints in the loss function. The physics-guided neural network (PgNN) embeds domain knowledge into the neural network by introducing the difference between the prediction results and the physical mechanism in the loss function \cite{daw2017physics}. On this basis, the physics informed neural network (PINN) was proposed \cite{PINN}, which can embed the governing equations, boundary conditions, and initial conditions into the neural network. In recent years, researchers have carried out a large amount of research on PINN, among which a typical application is to predict velocity and pressure fields in fluid mechanics based on PINN \cite{raissi2020hidden}. In order to utilize prior information, such as expert experience and engineering control, as domain knowledge in neural networks, \citeauthor{TgNN} [\citeyear{TgNN}] proposed the theory-guided neural network (TgNN) based on PINN and TGDS \cite{karpatne2017theory}. TgNN has achieved good performance in the field of hydrology and petroleum engineering. The computation time of the surrogate model of seepage process developed based on TgNN is only 10$\%$ of that of numerical simulation, reflecting the advantages of knowledge-embedded machine learning \cite{wang2021efficient}. \citeauthor{zhu2019physics} [\citeyear{zhu2019physics}] demonstrated that it is even possible to construct a loss function only based on domain knowledge and train a neural network without labeled data.

\begin{figure}
    \centering
    \includegraphics[width=8cm]{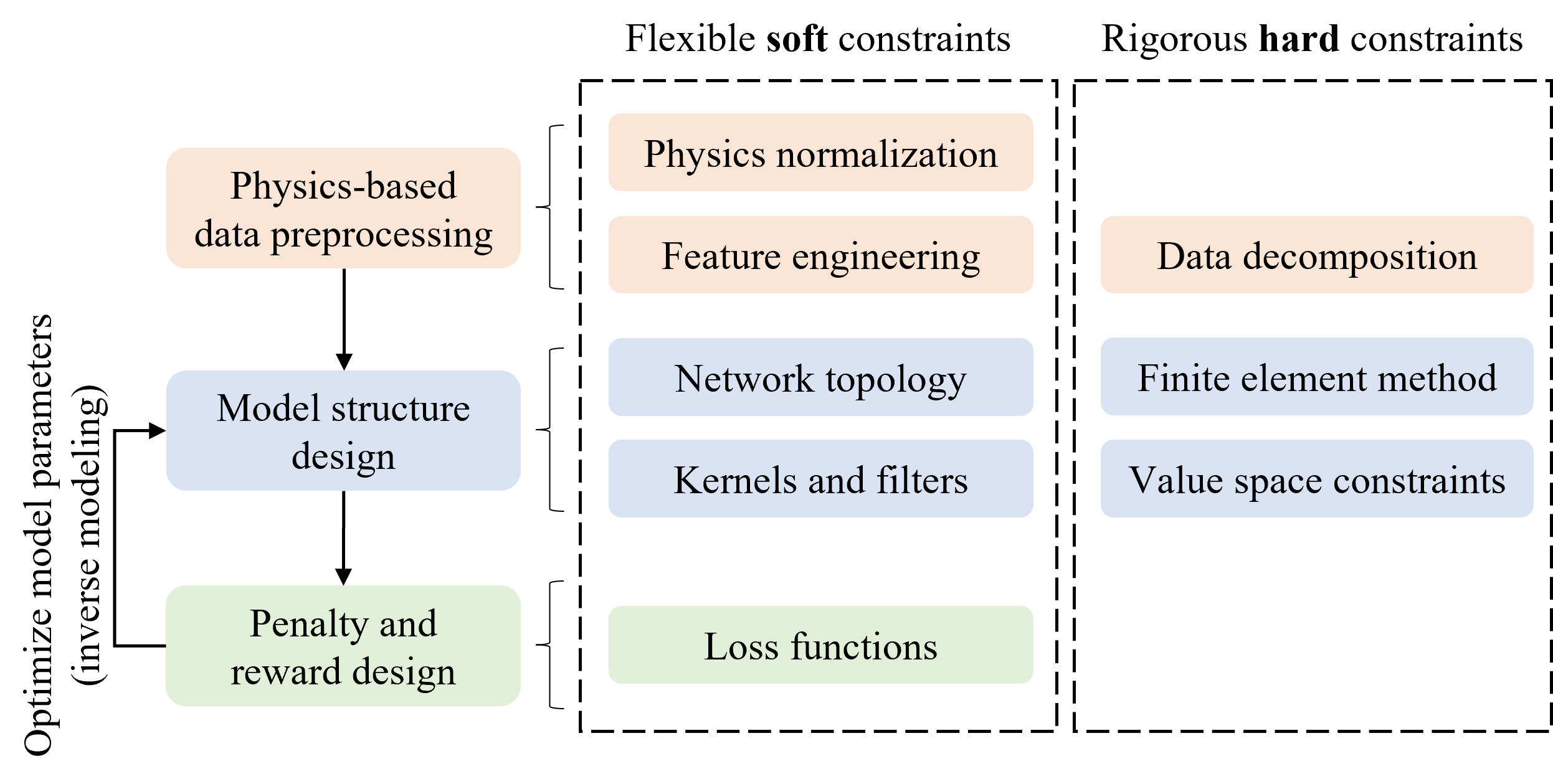}
    \caption{Diagram of the classification of knowledge embedding algorithms.}
\end{figure}

\subsection{Soft constraints and hard constraints}

In addition to analyzing knowledge embedding methods from the perspective of the machine learning modeling process, these methods may be separated into soft constraints and hard constraints from the standpoint of optimization. Soft constraints are easier to implement, while hard constraints ensure that the model outputs strictly adhere to known physical mechanisms.

Specifically, soft constraints introduce domain knowledge as prior information to the model, but do not require the model outputs to exactly comply with the domain knowledge. Figure 5 depicts the various types of soft constraints. The most typical soft constraint is to use the loss function to quantify the degree of consistency between the predictions and the physical mechanism. The domain knowledge can also be reflected through network topology or kernels and filters. Feature engineering and normalization are also used as soft constraints in the data preprocessing step. Although the soft constraints are easy to implement, they can only ensure that the predictions are close to the physical constraints (i.e., domain knowledge) on average, while they may generate predictions that violate the physical mechanism.

From an optimization perspective, hard constraints are more efficient methods than soft constraints, in general. Current studies on hard constraints in deep learning are still preliminary. \citeauthor{xu2022physics} [\citeyear{xu2022physics}] proposed physics constrained learning (PCL) to embed constraints into model by directly solving PDE. In the same year, \citeauthor{mohan2020embedding} [\citeyear{mohan2020embedding}] developed a physics-embedded decoder through the kernel of the convolutional neural network, and then embedded hard constraints in the neural network. \citeauthor{gao2021phygeonet} [\citeyear{gao2021phygeonet}] proposed to strengthen the initial conditions and Dirichlet boundary conditions by hardcoding in neural networks. Furthermore, value space constraints can also ensure that the outputs adhere precisely to the physical constraints \cite{HCP,luo2021deep}. Theoretically, since hard constraints can make better use of domain knowledge, the data requirements of the model can be reduced, and higher prediction accuracy can be obtained. However, because the hard constraint methods are highly dependent on the correctness of constraints, only accurate principles (e.g., law of conservation of energy) can be used as domain knowledge in practice.

\subsection{Challenges of knowledge embedding}

Domain knowledge essentially belongs to the semantic space, and machine learning models are in the vector space. Therefore, the core problem of knowledge embedding is to connect the semantic space and the feature space. At present, the challenges faced by knowledge embedding mainly include:

\begin{itemize}
\item The form of the embedded governing equations in existing models is simple and cannot handle complex scenarios. The complexity of the governing equations includes: 1. Existence of high-order derivatives or discontinuous data distribution, and the weak form of PDE might be a possible solution \cite{xu2021deep}; 2. Many constraints are inequalities and cannot be easily embedded into loss function, such as the engineering controls introduced by \citeauthor{wang2021efficient} [ \citeyear{wang2021efficient}]; 3. There may be source and sink terms in the equation; 4. The governing equations may be multiple coupled equations.
\item The basic models of knowledge embedding are mainly fully connected neural networks (for discrete sampling points) and convolutional neural networks (for regular physical fields). In fact, however, there are many irregular fields. The application of graph neural networks in knowledge embedding deserves further investigation.
\item The methods for inserting soft constraints into the loss function always contain many hyperparameters for regularization terms. The loss can be defined as $Loss=\sum_{n=1}^N\lambda_nl_n$, where $\lambda_n$ denotes hyperparameters and $l_n$ represents regularization terms. Different terms have different physical meanings and dimensions, and their impacts vary at different phases of optimization. Consequently, adaptive hyperparameters are worth exploring. 
\item Data in the real world are frequently scarce and noisy. In the future, strategies such as active learning, transfer learning, and employing neural networks to reduce noise \cite{PeRCNN,DLGA} should be investigated.
\item It is possible to make knowledge embedding models more accessible through auto machine learning and other methods, which enables engineers without a machine learning background to address actual issues.

\end{itemize}

\section{Discussion and Conclusion}

We systematically review studies on the integration of knowledge and data from the perspectives of knowledge discovery and knowledge embedding. This study evaluates and categorizes knowledge discovery algorithms based on the complexity of the structure and coefficients of the uncovered equations, as shown in Figure 2. In addition, this study summarizes the methods of embedding domain knowledge in the modeling process, and discusses the differences of soft constraints and hard constraints, as shown in Figure 5. 

Furthermore, we propose five research gaps and future opportunities for knowledge discovery and knowledge embedding, respectively. Suggestions for knowledge discovery include: building a more appropriate embedding approach for optimization with gradient-based methods; finding necessary conditions through multiple experiments; handling governing equations with both complex structures and complex coefficients; improving the accuracy of gradient computations; and simplifying equations found by symbolic mathematical methods. Regarding knowledge embedding, the research opportunities are: exploring approaches to embed complex governing equations; attempting to use network structures, such as graph neural networks, to handle irregular fields; implementing adaptive hyperparameters in soft constraints; focusing on noisy and scarce real-world data; and utilizing tools, such as auto machine learning, to lower the threshold for applying knowledge embedding models. Moreover, as illustrated in Figure 1, this study establishes a closed loop between knowledge discovery and knowledge embedding, realizing mutual promotion between domain knowledge (i.e., science) and machine learning models (i.e., engineering).

\section*{Acknowledgments}
This work is partially funded by the National Natural Science Foundation of China (Grant No. 62106116).

\bibliographystyle{named}
\bibliography{ijcai22}

\end{document}